\documentclass{article}

\usepackage[preprint]{neurips_2019}

\usepackage[utf8]{inputenc} 
\usepackage[T1]{fontenc}    
\usepackage{hyperref}       
\usepackage{url}            
\usepackage{booktabs}       
\usepackage{amsfonts}       
\usepackage{nicefrac}       
\usepackage{microtype}      

\usepackage{amsmath}
\DeclareMathOperator*{\argmax}{argmax}
\usepackage{pgfplots}
\usepackage{subcaption}
\usepackage{tabulary}
\usepackage{bm}
\newcolumntype{C}[1]{>{\centering\let\newline\\\arraybackslash\hspace{0pt}}m{#1}}
\newcolumntype{L}[1]{>{\let\newline\\\arraybackslash\hspace{0pt}}m{#1}}
\usepackage{dblfloatfix}
\usepackage{adjustbox}

\title{Continuous-action Reinforcement Learning for Playing Racing Games: Comparing SPG to PPO}

\author{
  Mario S. Holubar\\
  Department of Artificial Intelligence\\
  Bernoulli Institute\\
  University of Groningen\\
  \texttt{mario.holubar@gmail.com} \\
  \And
  Marco A. Wiering\\
  Department of Artificial Intelligence\\
  Bernoulli Institute\\
  University of Groningen\\
  \texttt{m.a.wiering@rug.nl} \\
}

\begin{document}

\maketitle

\begin{abstract}
In this paper, a novel racing environment for OpenAI Gym is introduced. This environment operates with continuous action- and state-spaces and requires agents to learn to control the acceleration and steering of a car while navigating a randomly generated racetrack. Different versions of two actor-critic learning algorithms are tested on this environment: Sampled Policy Gradient (SPG) and Proximal Policy Optimization (PPO). An extension of SPG is introduced that aims to improve learning performance by weighting action samples during the policy update step. The effect of using experience replay (ER) is also investigated. To this end, a modification to PPO is introduced that allows for training using old action samples by optimizing the actor in log space. Finally, a new technique for performing ER is tested that aims to improve learning speed without sacrificing performance by splitting the training into two parts, whereby networks are first trained using state transitions from the replay buffer, and then using only recent experiences. The results indicate that experience replay is not beneficial to PPO in continuous action spaces. The training of SPG seems to be more stable when actions are weighted. All versions of SPG outperform PPO when ER is used. The ER trick is effective at improving training speed on a computationally less intensive version of SPG.
\vspace{6ex}
\end{abstract}
\section{Introduction}\label{sec:introduction}

Reinforcement Learning (RL) (\cite{sutton2018reinforcement}) is an Artificial Intelligence paradigm which aims to develop policies for arbitrary tasks using a reward function as a supervision signal. By trying different actions in some environment and observing the outcome, an agent should be able to develop an idea of what to do in which situation in order to maximize the reward signal. A popular framework for this is actor-critic learning (\cite{konda2000actor}). This method uses two neural network function approximators, often called the actor and critic networks; the former selects actions to take in the environment, and the latter judges the quality of actions. As the actor interacts with the environment, the critic learns how its actions affect the reward signal. It can then teach the actor to perform better actions. The longer this process is repeated, the better both networks become at their tasks.

Because of the large data requirement and amount of trial and error necessary to learn a good policy, RL algorithms are usually trained in a simulated environment. This can be an abstract control task, a video game or a recreation of the real-world environment the agent will be deployed in.

In simulated environments, it is easy to provide the agent with a carefully selected set of features in order to maximize its performance. It is also possible to have the agent simply select one out of a list of predefined actions at each timestep, since the environment handles the actual execution of the actions. For robots to act in the real world however, it is important that they are able to understand the data produced by their sensors, which is almost always of continuous nature. They also often have to be able to exert precise control over their actuators in order to interact with their environment effectively. For these reasons, it is important to develop reinforcement learning algorithms that can deal with continuous state- and action-spaces.

In this paper, a racing environment for the OpenAI Gym (\cite{brockman2016openai}) baseline is introduced. In this task agents control a car and try to drive as far along a racetrack as they can, obtaining rewards based on their speed. In order to gain the highest reward possible, the agent has to learn to steer and accelerate or brake as necessary.

The main focus of this paper will be the actor-critic learning algorithms Sampled Policy Gradient (SPG) (\cite{wiehe2018sampled}) and Proximal Policy Optimization (PPO, \cite{schulman2017proximal}). SPG is an algorithm that updates the actor using action samples chosen by a Q-value network. It acts as a bridge between Continuous Actor Critic Learning Automaton (CACLA) (\cite{van2007reinforcement}) and Deterministic Policy Gradient (DPG) (\cite{silver2014deterministic}). Two different configurations of SPG will be tested on the racing environment and the effect of prioritizing action samples based on their Q-value will be investigated. SPG will also be compared to an implementation of PPO that does not use experience replay (ER), and a version of PPO with a modified objective function that is able to utilize ER.

\subsection{Contributions of this paper}
\begin{itemize}
    \item The racing environment is interesting for research since it acts as a simple baseline for continuous control, but can be extended in a large variety of ways. It also models a real-world task, making it particularly useful. For example, an agent could be pretrained using progressively more complex versions of the racing environment before being deployed in a real robot, minimizing the amount of trial and error required to reach good performance.
    \item SPG is compared to a state-of-the-art learning algorithm (PPO), a version of which was used in OpenAI's Dota 2 agent OpenAI Five\footnote{https://blog.openai.com/openai-five/}. This comparison is especially important since both of these approaches are fairly new and this comparison has not been made before.
    \item A modification of PPO is introduced that allows for the use of experience replay in a continuous-action environment.
    \item An extension to SPG is introduced that aims to improve training performance by weighting action samples during the actor update step.
    \item An extension to experience replay is introduced that aims to improve learning speed while retaining the advantages of full ER.
\end{itemize}

\subsection{Outline of the paper}
Section \ref{sec:rl} explains the background of the algorithms used in this paper. Section \ref{sec:methods} describes the environment, the PPO and SPG modifications, the experience replay trick and the experimental setup. The results are presented in section \ref{sec:results}, followed by their discussion in section \ref{sec:Conclusions}.
\section{Reinforcement learning}\label{sec:rl}

A reinforcement learning problem is generally modeled as a Markov Decision Process (MDP). An MDP is a process that takes an agent from one state to another, whereby the transition probabilities between different states depend only on the current state and the action the agent takes. For each state transition, the agent is given a reward $r_t$. The aim of the agent is to maximize the sum of future discounted rewards, also known as the gain (G), at every timestep t:
\begin{equation}
    G_t = \sum_{k=t}^\infty r_{k} \gamma^{k-t}
\end{equation}
$\gamma$ is the discount factor, which controls how rewards are weighed. A lower discount factor means that immediate rewards are preferred, while a discount factor close to 1 should be used for environments in which actions have long-lasting consequences.

\subsection{Actor-critic algorithms}
The algorithms used in this paper are similar to the advantage actor-critic (A2C) (\cite{konda2000actor}) learning method.
At the core of this approach are two neural networks: one called the actor (or policy network) and one called the critic (or value network). The actor takes the current state $s_t$ as its input and returns an action to be taken. This action $a_t$ is then passed to the environment, in which it is executed, producing a new state. This is repeated for a set number of steps or until the agent has reached its goal. This sequence is considered as one episode. During an episode, every timestep is assigned a reward $r_t$ depending on a reward function. The rewards are used by the critic to learn to estimate the gain for each step. The critic is therefore defined as:
\begin{equation}
    V^\pi(s) = \hat{\mathbb{E}}_\pi[G_t | s_t = s]
\end{equation}
The Monte-Carlo learning rule used to update the critic is:
\begin{equation}
    V^\pi(s_t) \leftarrow V^\pi(s_t) + \alpha (G_t - V^\pi(s_t)),
\end{equation}
where $\alpha$ is the learning rate.

The difference between the actual discounted reward for some state and the critic's prediction for it can be considered as an estimate of how much better/worse than expected it is. This value is known as the advantage $\hat{A}_t$, defined as:
\begin{equation}
    \hat{A}_t = G_t - V^\pi(s_t)
\end{equation}
The advantage can be used to train the actor so that actions that resulted in unexpectedly good outcomes are made more likely, and ones that resulted in worse outcomes are made less likely.

\subsection{Continuous action RL}
Many reinforcement learning algorithms such as Q-learning (\cite{watkins1992q}) use discrete action spaces, which is sufficient for most tasks that are used in research. However, in some applications it is not possible to discretize the action space, either because it would result in too many discrete choices, or because precise control is needed. This paper will discuss approaches that work in a continuous action space.

A central question in RL research is the exploration vs exploitation dilemma. In other words, to what degree should the agent try different things to learn more about its environment vs use its already acquired knowledge to perform as well as it can? This problem can be handled in various ways, such as $\epsilon$-greedy or softmax exploration (\cite{tijsma2016comparing}) (in the case of Q-learning) or by having the policy network output a probability for each action to be taken.

In order for the agent to act and explore in a continuous action space, Gaussian noise could be added to the output of the actor with a decreasing standard deviation, similarly to $\epsilon$-greedy exploration. In this paper however a different approach is used for both PPO and SPG, in which the policy network is given two output heads which correspond to the $\mu$ and $\sigma$ parameters of a normal distribution. The actions that are taken in the environment are sampled from this distribution. The policy is updated by minimizing or maximizing the log likelihood of the action being taken, depending on the advantage. The corresponding objective function, which needs to be maximized, is
\begin{equation}
    J^{PG}(\theta) = \hat{\mathbb{E}}_t [\log\pi_{\theta}(a_t|s_t)\hat{A}_t)].
\end{equation}
This allows for dynamic control over the exploration factor for each dimension of the action space separately. If accelerating is always correlated with higher reward, for example, the actor will reduce the standard deviation on the throttle output and focus on steering instead. If the continued acceleration then leads to crashes, the exploration on the throttle is increased again.

This paper will explore the effectiveness of two fairly recent actor-critic learning algorithms that are able to function in continuous spaces, which will be described now.

\subsection{PPO}
One problem of traditional actor-critic methods is that there is no guarantee for policy improvement; if the advantage at some state is negative, we only know that the action taken at that state should be made less likely, but not how much less likely. If multiple epochs of gradient descent are performed, it is easy for the gradient update to become too large, resulting in the policy being moved to an entirely unexplored (and potentially worse) area of the action space.

The idea of trust regions was introduced by \cite{schulman2015trust} to combat this problem. It guarantees an improving policy by constraining the size of the policy update based on the KL-divergence of the old and the current policy. This has been shown to improve performance, however it comes at the cost of simplicity. Trust Region Policy Optimization (TRPO) is not easy to implement and is also not compatible with certain network architectures.

This led to the introduction of Proximal Policy Optimization (\cite{schulman2017proximal}), which aims to combine the simplicity of vanilla policy gradient with the robustness and efficiency of TRPO. It does this by defining a probability ratio $p_t(\theta)$:
\begin{equation}
    p_t(\theta) = \frac{\pi_{\theta}(a_t|s_t)}{\pi_{\theta_{old}}(a_t|s_t)}
    \label{eq:rt}
\end{equation}
This ratio denotes the change the policy has gone through within an episode of training. $\pi_{\theta_{old}}$ is the policy at the beginning of the update. This means that during the first training epoch, $p_t(\theta) = 1$.

The clipped surrogate objective function $J^{CLIP}(\theta)$ is then maximized to update the policy:
\begin{equation}
J^{CLIP}(\theta) = \hat{\mathbb{E}}_t [min(clip(p_t(\theta), 1 - \epsilon, 1 + \epsilon)\hat{A}_t,p_t(\theta)\hat{A}_t)] + \beta H(\pi_{\theta}(s_t)),
\label{eq:PPO}
\end{equation}
where $\epsilon$ is a hyperparameter specific to PPO.

The entropy $H$ of a normal distribution is defined as:
\begin{equation}
    H = \frac{1}{2}ln(2 \pi e \sigma^2)
\end{equation}
Here, the entropy of the normal distribution given by $\pi_\theta(s_t)$ is multiplied by a factor $\beta$ and added to the objective to discourage premature convergence (\cite{williams1991function}). While this is not necessary, it was found to increase performance significantly.

The $\epsilon$ and $\beta$ values used in this research can be found in the appendix, along with all other hyperparameters discussed in this paper.

The clipped objective removes the incentive to move the policy far away from the old one, allowing for multiple epochs of optimization.
Taking the minimum of the clipped and unclipped terms results in the objective only being clipped if its value is improved by the new policy. If the value of the objective is worse under $\pi_\theta$ than under $\pi_{\theta_{old}}$, the update is allowed to be larger.

\subsection{SPG}
In Sampled Policy Gradient (\cite{wiehe2018sampled}) the critic, which is usually a state-value estimator, instead maps state-action pairs to Q-values. The term Q-value is used here because the critic architecture is the same as in Q-learning (\cite{watkins1992q}); it tries to predict the quality (Q) of a state-action combination:
\begin{equation}
    Q^\pi(s, a) = \hat{\mathbb{E}}_\pi[G_t | s_t = s, a_t = a]
\end{equation}
In the original SPG paper, temporal difference (TD) learning is used to update the critic in an off-policy way. TD-learning relies on a process called bootstrapping to update the critic with regard to an existing estimate of the gain $G_t$. This, combined with off-policy learning and function approximation, forms the "deadly triad" of reinforcement learning (\cite{sutton2018reinforcement}), which is known to cause instability and divergence of the critic. In this research, a variant of SPG that utilizes Monte-Carlo learning is used instead. The update rule for the critic is therefore:
\begin{equation}
    Q^\pi(s_t, a_t) \leftarrow Q^\pi(s_t, a_t) + \alpha (G_t - Q^\pi(s_t, a_t))
\end{equation}

In the policy update step, instead of performing gradient descent using the action that was taken and its corresponding advantage, actions $a_s$ are sampled from the action space. The critic is then used to obtain a Q-value for each of the actions. The sampled action with the highest Q-value is used as the backpropagation target for the gradient update of the actor:
\begin{equation}
    Target(s_t) = \argmax_{a_s} Q^{\pi}(s_t, a_s)
\end{equation}
The objective function for the actor that needs to be maximized is defined as:
\begin{equation}
    J^{SPG}(\theta) = \hat{\mathbb{E}}_t [\log\pi_{\theta}(Target(s_t)|s_t)].
    \label{eq:SPGUpdate}
\end{equation}

Several extensions to SPG are explained in the original paper. The algorithm also allows for any search strategy to be used for sampling actions. In this paper however, the standard algorithm is used with a simple Gaussian exploration strategy; actions are sampled around the taken action $a_t$ with an initial standard deviation $T$ that decays by some factor $\gamma_T$ after every episode. The original action $a_t$ is also included in the Q-value comparison.

SPG is comparable to Deterministic Policy Gradient (DPG) (\cite{silver2014deterministic}) since it uses the same critic architecture. However, whereas DPG updates the actor deterministically by 
taking the derivative of the value function towards the action and updating the actor based on this, SPG employs a more global search strategy. In theory, this makes it much less likely to get stuck in local optima.
\section{Methods}\label{sec:methods}

\subsection{Environment}\label{sec:environment}
\begin{figure}[b]
    \centering
    \includegraphics[width=0.4\textwidth]{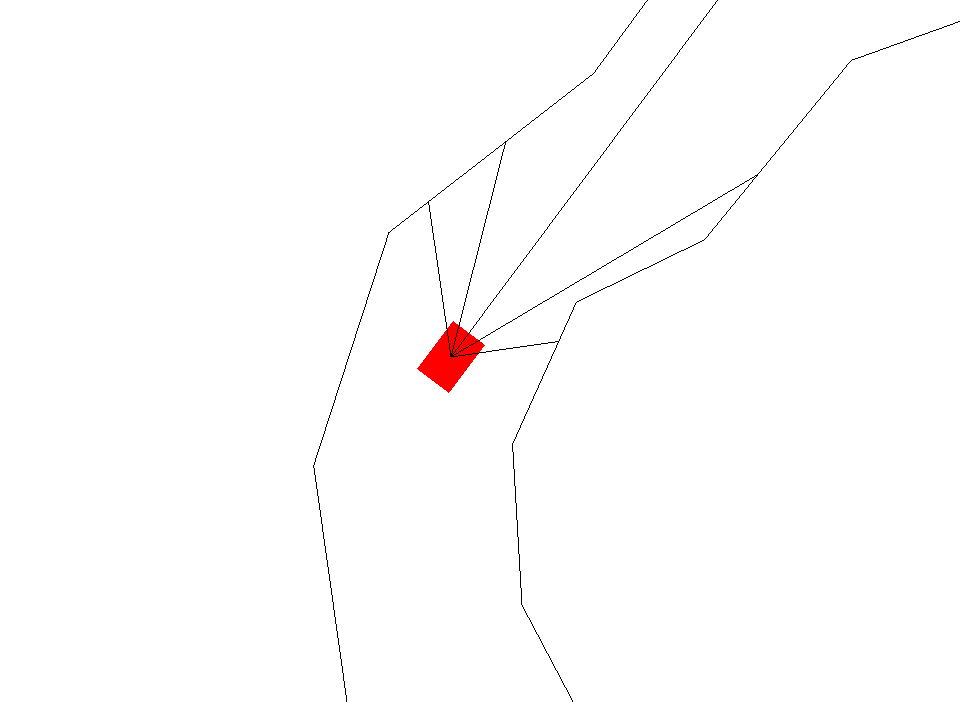}
    \caption{Screenshot of the environment depicting the agent in red with rays going towards the sides of the racetrack.}
    \label{fig:screenshot}
\end{figure}

For this research, we introduce a simple racing game environment, in which agents are represented as a car. They have to learn to accelerate, brake and steer as appropriate in order to navigate a racetrack.
The agent perceives its environment by the means of five distance sensors pointing away from the car at fixed relative angles. They measure the distance between the car and the side of the racetrack. The current speed is also known to the agent. This set of sensors has been shown to be optimal (\cite{togelius2007computational}). It also ensures that the only information that is utilized to make decisions is data that would be available to a real robot, e.g. in the form of LiDAR and motor sensors. Figure \ref{fig:screenshot} shows a visualization of the environment and the distance sensors.

The actions the agent can take at any given time consist of throttle control and steering. The former denotes the velocity the car is trying to reach; the motor will accelerate or brake as needed until it matches the desired speed. The latter controls the rotation of the car and is dependent on its velocity; the faster it is, the less it can turn. When the speed is low, there is a minimum turn radius that prevents it from making u-turns too quickly.

Both the states and actions are continuous and normalized between -1 and 1. The throttle control is remapped so that it outputs a value between 0 and the top speed of the car.

The racetrack is randomly generated and is made of a number of quadrilateral polygons that act as checkpoints. If the agent drives off the racetrack, the car is placed in the middle of the last passed checkpoint with a speed of zero.

The reward given to the agent every timestep is its current velocity (in pixels/sec) divided by its top speed. In order to gain the highest amount of total reward, the agent must learn to manage its speed to be as fast as possible, while avoiding crashing into walls. The reasoning behind this choice of reward metric is that it could also be measured by the agent in a real-world scenario and does not require the use of external tools or knowledge about the racetrack. Furthermore, a reward of -10 is given if the agent leaves the racetrack, however this is only necessary in this simulated environment where agents respawn immediately after crashing. A real robot would lose reward implicitly by being stuck on a wall; alternatively, another set of sensors could be used to detect collisions.

\subsection{Experience replay methods}
In order to use the information gained during learning as efficiently as possible, it should be used in multiple episodes of training. A method called experience replay (\cite{lin1993reinforcement}) is often used to allow this. The idea is that instead of only using the state transitions from the current episode in memory to train the agent, experiences are instead stored in a memory buffer. The buffer keeps experiences until it is full, at which point the oldest experiences are replaced. The agent's value and policy networks can be trained using this larger collection of data, allowing them to train in a more robust way without requiring more interactions with the environment.

An issue with this method is that agents are now slower to incorporate new information. For example, assume an agent has been training for a while. It has improved to the point where it now encounters a new obstacle that it has never seen before. With experience replay enabled, only a small fraction of the learning step will actually deal with this new problem, since the memory still consists of mostly old experiences. Only once the agent consistently encounters the problem for a while does it learn how to deal with it effectively. The result is slower training.

A potential solution to this issue is to split the training process into two parts: first, the agent is trained using experiences from the replay buffer. Then the same process is repeated with just "new" experiences, that is, ones that were obtained during the most recent episode. In theory, this allows the agent to retain the increased robustness gained from experience replay while also being able to react to new discoveries quickly.

\subsection{PPO with experience replay}
PPO is an on-policy learning algorithm, in which the optimizer uses actions that were taken under the current policy $\pi$. In theory, this prevents the use of experience replay (\cite{mnih2013playing}). However, in practice it has been shown that combining on-policy learning with experience replay is useful under certain conditions (\cite{sovrano2019combining}).

Just using the normal PPO implementation with experience replay does not work in a continuous-action environment. The reason for this is that to compute the loss value, PPO uses the probability ratio $p_t(\theta)$ as described in equation \ref{eq:rt}, working with the assumption that the action was taken under $\pi_{\theta_{old}}$. If this is not the case, and the parameters of the action sampling distribution have changed since the action was recorded, $\pi_{\theta_{old}}(a_t|s_t)$ can quickly tend towards 0, making $p_t(\theta)$ tend towards infinity. If the advantage associated with $a_t$ is negative, the ratio is not clipped, and the loss takes on an extraordinarily large value, leading to exploding gradients.

In order to be able to use PPO with experience replay, the solution we propose is to convert the algorithm to log space.

Let $\pi_{\theta}(a_t|s_t)$ equal the log likelihood of selecting an action $a_t$ at state $s_t$. The probability ratio is then defined as follows:
\begin{equation}
    p_t(\theta) = \pi_{\theta}(a_t|s_t) - \pi_{\theta_{old}}(a_t|s_t)
\end{equation}
Under this equation, $p_t(\theta_{old}) = 0$.

The final step is to modify the clipping operator. It is possible to use a hyperparameter $\epsilon_{log}$ for this and clip between $-\epsilon_{log}$ and $\epsilon_{log}$, but to allow for a fair comparison, the same hyperparameter $\epsilon$ is used as in default PPO and the objective is clipped between $log(1-\epsilon)$ and $log(1 + \epsilon)$. The full objective term that needs to be maximized then becomes:
\begin{equation}
    L^{CLIP}(\theta) = \hat{\mathbb{E}}_t [min(clip(p_t(\theta), log(1-\epsilon),log(1 + \epsilon))\hat{A}_t, p_t(\theta)\hat{A}_t)] + \beta H(\pi_{\theta}(s_t))
\end{equation}
The log-space transformation itself should not negatively affect PPO's performance, since $log(x)$ is monotonic with respect to x. In fact, gradient descent methods are generally better at optimizing a function in log space when dealing with probabilities, since the gradient of $log(p(x))$ is more well-scaled than the one of $p(x)$. Additionally, it is more numerically stable and not at risk of running into underflow problems.

\subsection{Prioritized SPG}
In most on-policy actor-critic algorithms such as PPO, each action is assigned an advantage based on its relative value. This advantage acts as a weight during the actor update step; it controls not only the direction but also the amount of the policy shift. For example, if an action led to a significantly worse outcome than expected, it is more important to make this action less likely to occur than if it only slightly worsened the agent's expected performance.

SPG does not have access to advantage values for sampled actions, and does not require their use: the estimated quality of the target action $Target(s_t)$ is always equal to or higher than that of the original action $a_t$, so it should never be made less likely. However, weighting the actions might still provide some benefit, and is made quite easy thanks to the Q-value estimation: the advantage of an action can be defined as the increase of its Q-value over that of the original action $a_t$. Therefore,
\begin{equation}
    \hat{A}_t = Q(s_t, Target(s_t)) - Q(s_t, a_t)
\end{equation}
and the new objective function becomes:
\begin{equation}
    J^{SPG-p}(\theta) = \hat{\mathbb{E}}_t [\log\pi_{\theta}(Target(s_t)|s_t))\hat{A}_t]
    \label{eq:SPGUpdateP}
\end{equation}
This way, sampled actions that only offer a slight increase in Q-value do not affect the policy as much, while ones that are estimated to be more useful are weighed more heavily.

The SPG variant that uses prioritization will be denoted as SPG-p.

\subsection{Experiment setup}
Due to the fact that SPG requires action samples to be evaluated by the critic for every step in an episode, the overhead associated with it is quite high. This is especially the case when the policy optimizer is run for multiple epochs. Hence, two versions of SPG with different hyperparameters will be compared: one that uses the same learning hyperparameters as PPO, meaning the actor update step is essentially run multiple times in a row with a low learning rate; and one in which the update is only performed once per episode, with a higher learning rate.

These two SPG configurations will be tested three times each: once without experience replay, so the actor and critic are updated using only the most recent episode's experiences; once with purely experience replay, meaning that after an episode the collected experiences are added to the replay buffer and then the networks are trained using the whole buffer; and once with the hybrid approach in which the networks are first trained using only the replay buffer, and then using new experiences. In this case, both steps use half the amount of experiences so the total amount of state transitions given to the optimizer stays constant.

Additionally, the original PPO implementation using only recent experiences will be compared to the logarithmic version with and without experience replay, as well as to the different SPG variants.

An entropy bonus that encourages exploration was added to PPO's loss function since performance suffered significantly without it. It was determined in prior experiments that this entropy bonus does not negatively affect PPO's final performance. SPG did not have any issues of this sort. In fact it performed slightly worse with the entropy bonus, hence it was omitted.

\begin{figure}
    \centering
\begin{adjustbox}{width=\textwidth}
    \includegraphics[width=0.20\textwidth]{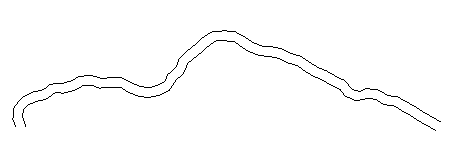}
    \includegraphics[width=0.08\textwidth]{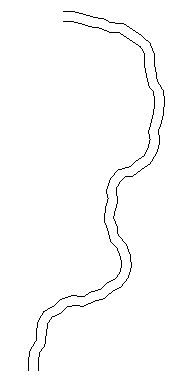}
    \includegraphics[width=0.20\textwidth]{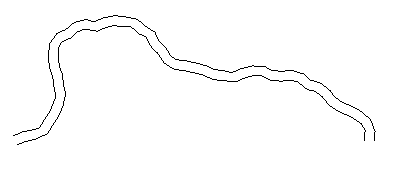}
    \includegraphics[width=0.12\textwidth]{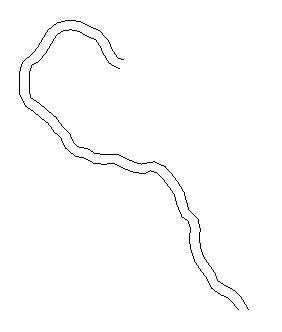}
    \includegraphics[width=0.08\textwidth]{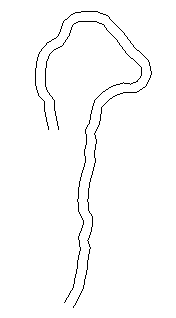}
\end{adjustbox}
    \caption{The five random racetracks used for the experiments.}
    \label{fig:tracks}
\end{figure}

All the experiments are repeated five times, where each time a different racetrack is used. The different seeds used to generate the racetracks are the same across algorithms to preserve fairness. Figure \ref{fig:tracks} shows the racetracks produced by these seeds.

Multilayer perceptrons (MLPs) are used as function approximators for the actor and critic. Their architecture is the same in all algorithms. The critic network consists of two hidden layers with 100 neurons each and the actor has one hidden layer with 100 neurons. The activation function of the hidden layers is tanh. No activation function is applied to the critic's output layer. The actor uses tanh and softplus for the $\mu$ and $\sigma$ output heads, respectively.
The critic and actor are optimized independently of each other after every episode. This means a different number of epochs can be used for the value and policy networks.
The critic is optimized first so that new experiences are already incorporated when it is used by SPG to sample actions during the actor update.
Every epoch, a minibatch of data is sampled from the experiences of the most recent episode or the replay buffer and used to perform gradient descent on the networks.

All hyperparameters used for this research can be found in the appendix. They were initially selected according to the original papers' recommendations and then tuned in preliminary experiments.

\section{Results}\label{sec:results}

All reward curves are produced by recording the reward obtained by the agent in each episode and averaging the results over the five racetracks.

\begin{figure}[h]
\centering
\begin{minipage}{0.48\linewidth}
    \centering
    \includegraphics[width=\linewidth]{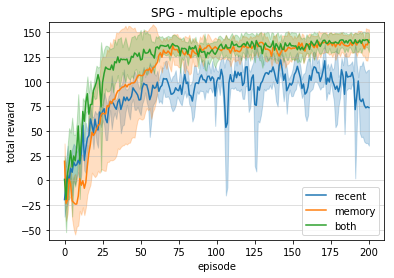}
    \caption{SPG reward curves with multiple training epochs.}
    \label{fig:spg-multi}
\end{minipage}\quad
\begin{minipage}{0.48\linewidth}
    \centering
    \includegraphics[width=\linewidth]{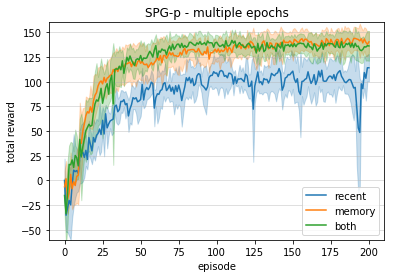}
    \caption{SPG reward curves with multiple training epochs and prioritization.}
    \label{fig:spgp-multi}
\end{minipage}
\end{figure}

\begin{figure}[h]
\centering
\begin{minipage}{0.48\linewidth}
    \centering
    \includegraphics[width=\linewidth]{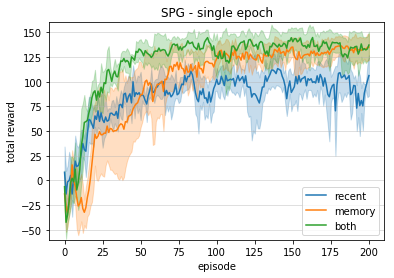}
    \caption{SPG reward curves with one training epoch.}
    \label{fig:spg-single}
\end{minipage}\quad
\begin{minipage}{0.48\linewidth}
    \centering
    \includegraphics[width=\linewidth]{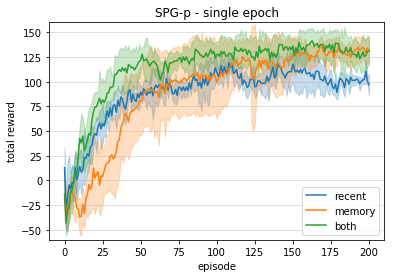}
    \caption{SPG reward curves with one training epoch and prioritization.}
    \label{fig:spgp-single}
\end{minipage}
\end{figure}

It is evident from figures \ref{fig:spg-multi} to \ref{fig:spgp-single} that weighting actions based on their Q-values does not have much effect on training speed or final performance; the learning curves of the two variants are very similar, whether one or multiple epochs of training is used. However, it does seem like training is slightly more stable with prioritization. This is characterized by fewer downward spikes in the learning curves and overall lower variance.

Since the difference between the results of these two variants is so small, only SPG-p is used for all other comparisons.

Looking at the performance of just the multiple epoch variants of SPG (figures \ref{fig:spg-multi}, \ref{fig:spgp-multi}), we can observe that experience replay gives it a large advantage. Training is more reliable and results in higher final performances. The hybrid variant that trains on both recent and old experiences seems to have equal or even better performance.

In figures \ref{fig:spg-single} and \ref{fig:spgp-single} the learning curves of the faster variant of SPG that trains using only one epoch each episode are shown. When experience replay is used, the final performance here is close to the one reached by the slower method. However, the learning process is initially much slower and only overtakes the recent-only performance when run for a sufficient number of episodes. Here, the algorithm benefits from the hybrid method, which seems to combine the learning speed of using recent experiences with the good final performance of experience replay.

\begin{figure}[h]
\centering
\begin{minipage}{0.48\linewidth}
    \centering
    \includegraphics[width=\linewidth]{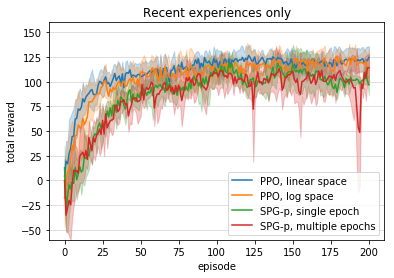}
    \caption{Reward curves of different approaches when no experience replay is used.}
    \label{fig:recent}
\end{minipage}\quad
\begin{minipage}{0.48\linewidth}
    \centering
    \includegraphics[width=\linewidth]{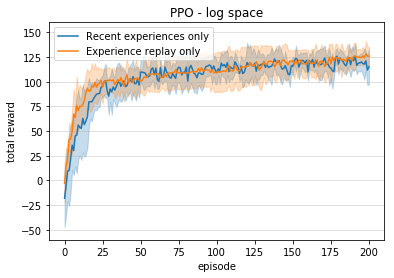}
    \caption{Log-space PPO reward curves with and without experience replay.}
    \label{fig:ppo_log}
\end{minipage}
\end{figure}

\begin{figure}[h]
\centering
\begin{minipage}{0.48\linewidth}
    \centering
    \includegraphics[width=\linewidth]{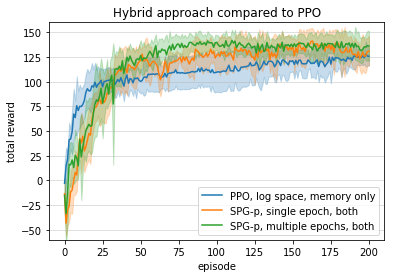}
    \caption{Reward curves of SPG variants using both recent and old experiences compared to PPO.}
    \label{fig:both}
\end{minipage}
\end{figure}

Figure \ref{fig:recent} shows that the linear space and log space versions of PPO have nearly identical performance when no experience replay is used. When SPG is used without experience replay, it is outperformed by PPO in both configurations. The multiple-epoch version of SPG offers no improvement over the faster version in this case. It even seems to be less stable, with some trials suffering from occasional temporary drops into negative reward, characterized by large downward spikes of the reward curve.

From figure \ref{fig:ppo_log} it is evident that experience replay does not improve the performance of PPO much. The variant with half new and half old experiences had issues with exploding gradients, despite using the log space. It was not able to run for 200 episodes without this problem occurring, and was therefore omitted from the results.

Figure \ref{fig:both} shows that the faster SPG method using the hybrid approach matches the performance of the slower multiple epoch variant. Both outperform PPO from episode 30 onwards.

\def\arraystretch{1.5}
\begin{table*}[h]
    \centering
    \begin{adjustbox}{width=\textwidth}
    \begin{tabulary}{\textwidth}{|C{1.3cm}|C{1.65cm}|C{1.65cm}|C{1.65cm}|C{1.65cm}|C{1.65cm}|C{1.9cm}|}
    \hline
     & \bf{PPO, linear} & \bf{PPO,\newline log} & \bf{SPG, single} & \bf{SPG, multiple} & \bf{SPG-p, single} & \bf{SPG-p, multiple}\\
    \hline
    \bf{Recent} & $121.5 \pm 5.8$ & $119.5 \pm 4.6$ & $94.4 \pm 3.8$ & $91.9 \pm 7.5$ & $101.2 \pm 2.6$ & $95.8 \pm 4.7$\\
    \hline
    \bf{Memory} &  & $124.2 \pm 4.9$ & $133.5 \pm 5.7$ & $138.2 \pm 4.9$ & $130.6 \pm 5.2$ & \bm{$141.1 \pm 5.7$}\\
    \hline
    \bf{Both} &  &  & $132.1 \pm 6.7$ & $140.9 \pm 4.0$ & $129.1 \pm 6.9$ & $135.9 \pm 6.2$\\
    \hline
    \end{tabulary}
    \end{adjustbox}
    \caption{Final performance of all configurations with standard error values. The rewards of the last 20 episodes of training are averaged to obtain these values.}
    \label{tab:all_results}
\end{table*}

Table \ref{tab:all_results} outlines the performance of all setups at the end of training. PPO outperforms SPG when only recent experiences are available, but it is outperformed by all SPG variants that use experience replay.
\section{Discussion}\label{sec:Conclusions}
The results show that while PPO is very proficient at incorporating new information, its potential is still hindered; 
SPG benefits much more from the use of experience replay, making it a promising algorithm for continuous-action reinforcement learning.

Prioritizing action samples in SPG seems to lead to increased stability of the learning process. This can be attributed to the fact that the policy is only changed significantly if an action sample is considered to have high relative quality.

The downward spikes of SPG's multiple-epoch performance when only recent experiences are used might be caused by the critic wrongly classifying certain places in the state-action space to have very high values; the actor may be better at finding these places due to the increased number of total action samples. The problem does not occur when experience replay is used because the critic is more stable thanks to the increased amount of training data.

Figure \ref{fig:ppo_log} outlines an important issue with on-policy experience replay. It does not seem to improve the performance of PPO much, if at all. It is possible that this issue occurs because of the log-space implementation, however the highly similar training curves in figure \ref{fig:recent} indicate that log-PPO's performance is representative of linear-PPO's performance. Since \cite{sovrano2019combining} showed that PPO can benefit from ER in a discrete action environment, it stands to reason that the stagnation of ER performance in this research has to do with the continuous-action variant of PPO. 

When SPG is given more samples to train on (e.g. using experience replay), the value network gets more and more accurate. The policy network improves as well as a direct result of this, since it is trained using samples chosen by the critic. In PPO, on the other hand, the policy network can only be trained using actions that have been taken in the past. As both the actor and the critic learn, more and more of the old actions will be considered as "bad" and assigned a negative advantage. This means that instead of the policy moving towards the best possible action, like in SPG, it only moves away from bad actions, which does not guarantee policy improvement.

As the policy changes, the old action samples not only become relatively worse, but also become less likely to be chosen by the current policy. When the probability of choosing an action $\pi_{\theta_{old}}(a_t|s_t)$ is extremely low, the probability ratio becomes very large. Combined with a negative advantage, this causes a high loss value, which moves the neural network weights by a large amount. This makes the action even more unlikely in the next epoch, leading to a snowball effect that causes exploding gradients. While the effect is reduced significantly by using log space, allowing for the use of experience replay, the variant using alternating old and new experiences still suffered from this problem. This can be attributed to the old and new actions pushing the gradients in different directions, making each other more and more unlikely.

The core issue preventing efficient use of PPO with experience replay in a continuous-action environment therefore seems to be twofold: firstly, actions taken under a different policy are not able to update the current policy efficiently; and secondly, actions with a very low likelihood of being chosen under the current policy affect the gradients more because they lead to a large loss value.

As shown in table \ref{tab:all_results}, using multiple epochs of gradient descent on SPG seems to yield an improvement in learning capability. It is not hindered by large policy updates in the way that a vanilla actor-critic implementation would be. This can likely be attributed to the action sampling; instead of moving towards or away from the same actions multiple times, each epoch uses different samples from the action space, which ultimately converge to a more global solution. This does come at the cost of increased computation overhead however.

In situations where computation time is critical, the single-epoch variant of SPG can be used. It is able to achieve a high performance when used with experience replay, but only after a large number of episodes. Here, the trick of using half old and half new experiences to train the actor seems to increase learning speed significantly. Figure \ref{fig:both} shows that when this tweak is applied, the learning curve is very similar to the one of the slower, more accurate SPG variant. It is able to outperform PPO relatively quickly this way.

Despite this, computation time is usually not the limiting factor of reinforcement learning and the trick does not seem to help much when used on the higher-performance SPG variant. However, this might be a problem with the task itself; the aim of the trick is to facilitate overcoming new obstacles that require adaptation of the policy. But the nature of the racing task is such that once the agent has learned to steer and slow down when it needs to, there is not much left to learn. The rest of the training is then only focused on tweaking the policy slightly, which the trick is not very useful for. 

It would be interesting to research if Prioritized Experience Replay (\cite{schaul2015prioritized}) would improve on the currently used ER variants.
Additionally, the performance of SPG could be investigated when compared to other off-policy methods that allow for continuous action spaces, such as NAF (\cite{gu2016continuous}), Trust-PCL (\cite{nachum2017trust}), or TD3 (\cite{fujimoto2018addressing}).

\enlargethispage{2.5\baselineskip}

\bibliographystyle{plainnat}
\bibliography{literature}

\clearpage
\section*{Appendix}\label{sec:Appendix}

\subsection*{Hyperparameters}

\def\arraystretch{1}

\subsubsection*{General}
\begin{tabular}{L{5.5cm}|L{2cm}}
    Steps per episode & 200\\
    ER buffer size (state transitions) & 10000\\
    Discount factor $\gamma$ & 0.9\\
    Frame skip & 0\\
\end{tabular}

\subsubsection*{Network setup (all algorithms)}
\begin{tabular}{L{5.5cm}|L{2cm}}
    \# Critic hidden layers & 2 \\
    \# Actor hidden layers & 1 \\
    \# Neurons per hidden layer & 100 \\
    Hidden layer activation function & tanh \\
    Optimizer & Adam \\
\end{tabular}

\subsubsection*{PPO, both versions}
\begin{tabular}{L{5.5cm}|L{2cm}}
    Critic learning rate & 0.0005 \\
    Actor learning rate & 0.001 \\
    Entropy factor $\beta$ & 0.02 \\
    Epsilon $\epsilon$ & 0.2 \\
    \# Value epochs & 50 \\
    \# Policy epochs & 10 \\
    Minibatch size & 200 \\
\end{tabular}

\subsubsection*{SPG, one epoch}
\begin{tabular}{L{5.5cm}|L{2cm}}
    Critic learning rate & 0.0005 \\
    Actor learning rate & 0.01 \\
    Entropy factor $\beta$ & 0.0 \\
    \# Action samples & 5 \\
    Initial exploration temperature $T$ & 1.0 \\
    Exploration temperature decay $\gamma_T$ & 0.01 \\
    \# Value epochs & 50 \\
    \# Policy epochs & 1 \\
    Minibatch size & 200 \\
\end{tabular}

\subsubsection*{SPG, multiple epochs}
\begin{tabular}{L{5.5cm}|L{2cm}}
    Critic learning rate & 0.0005 \\
    Actor learning rate & 0.001 \\
    Entropy factor $\beta$ & 0.0 \\
    \# Action samples & 5 \\
    Initial exploration temperature $T$ & 1.0 \\
    Exploration temperature decay $\gamma_T$ & 0.01 \\
    \# Value epochs & 50 \\
    \# Policy epochs & 10 \\
    Minibatch size & 200 \\
\end{tabular}

\subsection*{Code}
The code for this paper, including the racing task and SPG and PPO implementations as well as the results obtained can be found at https://github.com/mario-holubar/RacingRL.

\end{document}